\documentclass[graybox]{svmult}

\usepackage{mathptmx}       %
\usepackage{helvet}         %
\usepackage{courier}        %
\usepackage{type1cm}        %
\usepackage{makeidx}         %
\usepackage{graphicx}        %
\usepackage{multicol}        %
\usepackage[bottom]{footmisc}%

\usepackage[numbers]{natbib}
\usepackage{multicol}
\usepackage{graphicx}
\usepackage{enumerate}
\usepackage{subfig}
\usepackage{wrapfig}
\usepackage{sidecap}
\usepackage{floatrow}
\usepackage{amssymb} 
\usepackage{amsmath}
\usepackage{bbm}
\usepackage[font={footnotesize}]{caption}

\usepackage[usenames,dvipsnames,svgnames]{xcolor}

\usepackage{titlesec}
\titlespacing\section{0pt}{4pt plus 4pt minus 2pt}{2pt plus 0pt minus 2pt}
\titlespacing\subsection{0pt}{4pt plus 4pt minus 2pt}{2pt plus 0pt minus 2pt}
\titlespacing\subsubsection{0pt}{4pt plus 4pt minus 2pt}{2pt plus 0pt minus 2pt}

\usepackage{titlesec}
\titlespacing*{\subsection}{0pt}
    {1\baselineskip plus 0.3\baselineskip minus 0.38\baselineskip}
    {1\baselineskip plus 0.3\baselineskip minus 0.38\baselineskip}
\titlespacing*{\section}{0pt}
    {1\baselineskip plus 0.3\baselineskip minus 0.38\baselineskip}
    {1\baselineskip plus 0.3\baselineskip minus 0.38\baselineskip}

\makeindex             %

\DeclareMathOperator*{\minimize}{minimize}

\newcommand{\xr}{x_R}
\newcommand{\yr}{y_R}
\newcommand{\psir}{\psi_R}
\newcommand{\Uxr}{U_{x_R}}
\newcommand{\Uyr}{U_{y_R}}
\newcommand{\rr}{r_R}

\newcommand{\dUxr}{\dot{U}_{x_R}}
\newcommand{\dUyr}{\dot{U}_{y_R}}
\newcommand{\drr}{\dot{r}_R}

\newcommand{\Fx}{F_x}
\newcommand{\Fxf}{F_{x_f}}
\newcommand{\Fxr}{F_{x_r}}
\newcommand{\Fyf}{F_{y_f}}
\newcommand{\Fyr}{F_{y_r}}

\newcommand{\xh}{x_H}
\newcommand{\yh}{y_H}
\newcommand{\psih}{\psi_H}
\newcommand{\vh}{v_H}

\newcommand{\dvh}{\dot{v}_H}

\newcommand{\rel}{\mathcal{R}}
\newcommand{\xrel}{x_{rel}}
\newcommand{\dxrel}{\dot{x}_{rel}}
\newcommand{\yrel}{y_{rel}}
\newcommand{\dyrel}{\dot{y}_{rel}}
\newcommand{\psirel}{\psi_{rel}}
\newcommand{\dpsirel}{\dot{\psi}_{rel}}

\usepackage{url}

\begin{document}
\title*{On Infusing Reachability-Based Safety Assurance within Probabilistic Planning Frameworks for Human-Robot Vehicle Interactions}
\titlerunning{Reachability-Based Safety Assurance for Human-Robot Vehicle Interactions}
\author{Karen Leung$^*$, Edward Schmerling$^*$, Mo Chen,
John Talbot, J. Christian Gerdes,
and Marco Pavone}
\authorrunning{K. Leung, E. Schmerling, M. Chen, J. Talbot, J. C. Gerdes, and M. Pavone}
\institute{Karen Leung$^1$, Edward Schmerling$^2$,
John Talbot$^3$, J. Christian Gerdes$^3$,
Marco Pavone$^1$ \at Department of Aeronautics and Astronautics$^1$, Institute for Computational and Mathematical Engineering$^2$,
and Department of Mechanical Engineering$^3$,
Stanford University, Stanford, CA 94305, USA.
\email{{karenl7, schmrlng, john.talbot, cgerdes,
pavone}@stanford.edu}
\and Mo Chen, \email{mochen@cs.sfu.ca} \at
School of Computing Science, Simon Fraser University, Burnaby, BC V5A 1S6, Canada. \at
Work performed as a postdoctoral scholar at Stanford University.
\and $^*$Karen Leung and Edward Schmerling contributed equally to this work. This work was supported by the Office of Naval Research (Grant N00014-17-1-2433), by Qualcomm, and by the Toyota Research Institute (``TRI''). This article solely reflects the opinions and conclusions of its authors and not ONR, Qualcomm, TRI, or any other Toyota entity. The authors would like to thank Thunderhill Raceway Park for accommodating testing.}
\maketitle
\vspace{-2cm}
\abstract{
Action anticipation, intent prediction, and proactive behavior are all desirable characteristics for autonomous driving policies in interactive scenarios. Paramount, however, is ensuring safety on the road --- a key challenge in doing so is accounting for uncertainty in human driver actions without unduly impacting planner performance.
This paper introduces a minimally-interventional safety controller operating within an autonomous vehicle control stack with the role of ensuring collision-free interaction with an externally controlled (e.g., human-driven) counterpart. We leverage reachability analysis to construct a real-time (100Hz) controller that serves the dual role of (1) tracking an input trajectory from a higher-level planning algorithm using model predictive control, and (2) assuring safety through maintaining the availability of a collision-free escape maneuver as a persistent constraint regardless of whatever future actions the other car takes. A full-scale steer-by-wire platform is used to conduct traffic weaving experiments wherein the two cars, initially side-by-side, must swap lanes in a limited amount of time and distance, emulating cars merging onto/off of a highway. We demonstrate that, with our control stack, the autonomous vehicle is able to avoid collision even when the other car defies the planner's expectations and takes dangerous actions, either carelessly or with the intent to collide, and otherwise deviates minimally from the planned trajectory to the extent required to maintain safety.}

\section{Introduction}\label{sec:intro}

Decision-making and control for mobile robots is typically stratified into levels.
A high-level planner, informed by representative dynamics of a robot and its environment, might be responsible for selecting a coarse trajectory plan, which is implemented through a low-level controller that respects more accurate models of the robot's dynamics and control constraints.
While additional components may be required to flesh out a robot's full control stack from model to motor commands, selecting the right ``division of responsibilities'' is fundamental to system design.

One consideration that defies clear classification, however, is how to ensure a mobile robot's safety when operating in close proximity with a rapidly evolving and stochastic environment.
Safety is a function of uncertainty in both the robot's dynamics and those of its surroundings; high-level planners typically do not replan sufficiently rapidly to ensure split-second reactivity to threats, yet low-level controllers are typically too short-sighted to ensure safety beyond their local horizon. Hand-designed artificial potential fields have been employed at the planning level \cite{WolfBurdick2008} to encourage the selection of safer nominal trajectories, while reactive collision avoidance techniques, e.g., based on online optimal control \cite{FunkeBrownEtAl2017} or precomputed emergency maneuver libraries \cite{AroraChoudhuryEtAl2015}, have been applied at the controller level to avoid previously unseen static obstacles, but these approaches do not account for interactive scenarios in which another sentient agent is a key environmental consideration.

In this work we implement a control stack for a full-scale autonomous car (the ``robot'') engaging in close proximity interactions with a human-controlled vehicle (the ``human''). Freedom of motion is essential for the planner to carry out the driving task while conveying future intent to the other vehicle. Our primary tool for designing a controller that does not needlessly impinge upon the planner's choices is Hamilton-Jacobi (HJ) reachability.  Backward HJ reachability, in particular, has been studied extensively and applied successfully in a variety of safety-critical interactive settings \cite{BokanowskiForcadelEtAl2010,ChenHuEtAl2017,DhinakaranChenEtAl2017,GattamiAlAlamEtAl2011,MargellosLygeros2011,MitchellBayenEtAl2005} due to its flexibility with respect to system dynamics, and its optimal (i.e., non-overly-conservative) avoidance maneuvers stemming from its equivalence to an exhaustive search over joint system dynamics.
Previously reachability-based controllers have largely been used in a switched fashion: the robot is normally allowed to apply any control, but switches to the optimal avoidance controller when near safety violation.
While this is sometimes effective \cite{FisacAkametaluEtAl2017}, to the best of our knowledge there has not been any work that explicitly addresses the integration of reachability-based safety controllers as a component within a robot's control stack, i.e., with safety as a constraint upon a primary planning objective.

\noindent \textit{\textbf{Statement of Contributions:}} The contributions of this paper are twofold. First, we propose a method for formally incorporating reachability-based safety within an existing optimization-based control framework. The main insight that enables our approach is the recognition that, near safety violation, the set of safety-preserving controls often contains more than just the optimal avoidance control. Instead of directly applying this optimal avoidance control when prompted by reachability considerations, as in a switching control approach, we quantify the set of safety-preserving controls and pass it to the broader control framework as a constraint. Our intent is to enable truly minimal intervention against the direction of a higher-level planner when evasive action is required. Second, we evaluate the benefits and performance of this safe control methodology in the context of a probabilistic planning framework for the traffic weaving scenario studied at a high level in \cite{SchmerlingLeungEtAl2018}, wherein two cars, initially side-by-side, must swap lanes in a limited amount of time and distance. Experiments with a full-scale steer-by-wire vehicle reveal that our combined control stack achieves better safety than applying a tracking controller alone to the planner output, and smoother operation (with similar safety) compared with a switching control scheme; in our discussion we provide a roadmap towards improving the level of safety assurance in the face of practical considerations, i.e., unmodeled dynamics, as well as towards generalizations of the basic traffic weaving scenario.

\section{Technical Approach}\label{sec:approach}

\begin{figure}[t]
    \centering
    \includegraphics[width=\textwidth]{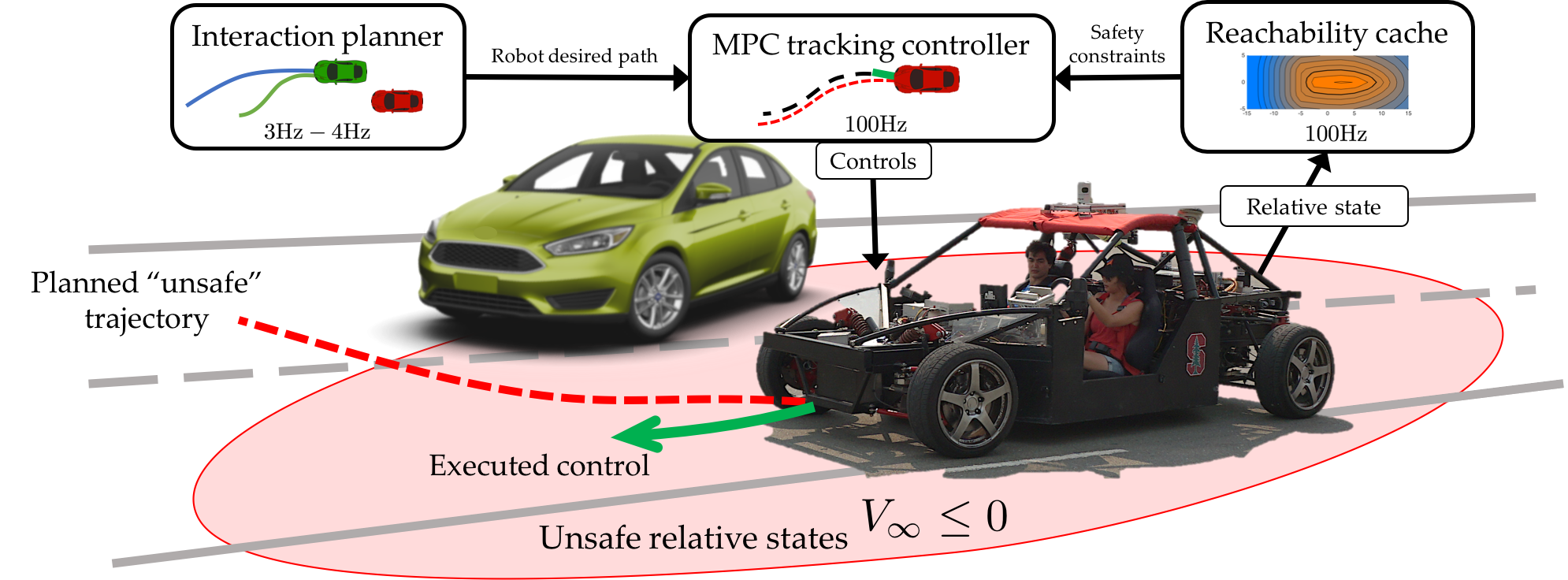}
    \caption{Decision-making and control stack for human-robot pairwise vehicle interactions.
    Our contribution in this work is the integration of safety-ensuring control constraints, derived from a HJ backward reachable set computed and cached offline, into a model predictive controller's tracking optimization problem. A high-level interaction planner produces nominal trajectories for the robot car (foreground); the low-level safe tracking controller executes controls that minimally deviate from the planner's choice if the vehicles approach the set of unsafe relative states.
    }
    \label{fig:schematic}
\end{figure}
To bring safe traffic weaving to the road, we combine a model-based interaction planner \cite{SchmerlingLeungEtAl2018} with a real-time trajectory tracking model predictive control (MPC) controller inspired by \cite{BrownFunkeEtAl2017}, modified to include an additional invariant set constraint detailed in this section. An outline of our control framework is shown in Figure~\ref{fig:schematic}. 
\noindent \emph{Interaction Planner: }The traffic-weaving interaction planner \cite{SchmerlingLeungEtAl2018} uses a predictive model of future human behavior to select a desired trajectory for the robot car to follow, updated at $\sim$3Hz. We extend this work by using a hindsight optimization policy \cite{YoonFernEtAl2008} instead of the limited-lookahead action policy used in \cite{SchmerlingLeungEtAl2018} in order to encourage more information-seeking actions from the robot.
Though this interaction planner optimizes an objective that weighs safety considerations (e.g., distance between cars) relative to other concerns (e.g., control effort), it reasons anticipatively with respect to a \emph{probabilistic} interaction dynamics model. That is, safety is not enforced as a deterministic constraint at the planning level.

\noindent \emph{MPC Tracking Controller: }The MPC tracking controller, operating at 100Hz, computes optimal controls to track a desired trajectory by solving a quadratic program (QP) at each iteration. 
This optimization problem is based on a single track vehicle model (also known as the bicycle model) and incorporates friction and stability control constraints while minimizing a combination of tracking error and steering effort.
We incorporate additional constraints computed from HJ reachability analysis into this QP to ensure that at each control step the robot car does not enter a state of possible inevitable collision.

\begin{figure}
\centering
\subfloat[\footnotesize Joint dynamics are propagated \emph{backwards} in time. The sets are not overly-conservative even for long time horizons. However, they are computationally expensive to compute.]{
    \label{fig:BRS illustration}
    \includegraphics[width=0.6\textwidth]{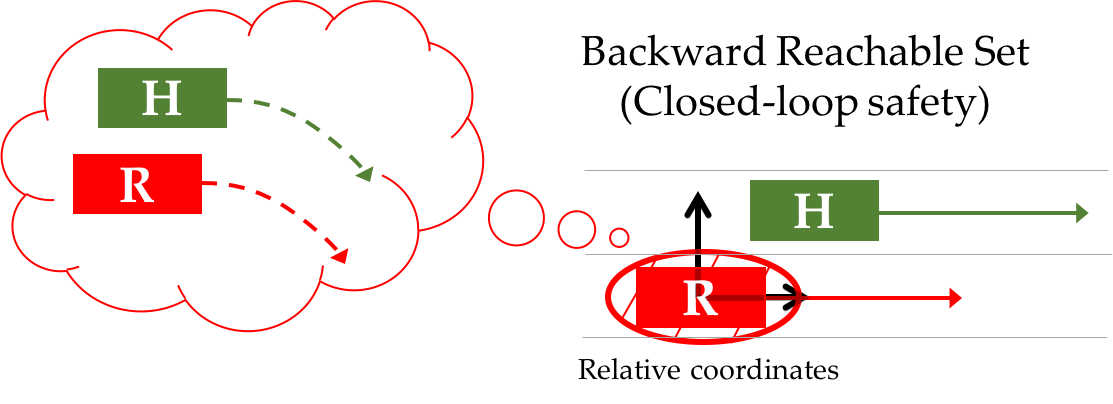}} 
\quad
\subfloat[\footnotesize Human dynamics are propagated \emph{forward} in time (overly-conservative in general).]{ %
    \label{fig:FRS illustration}
    \includegraphics[width=0.32\textwidth]{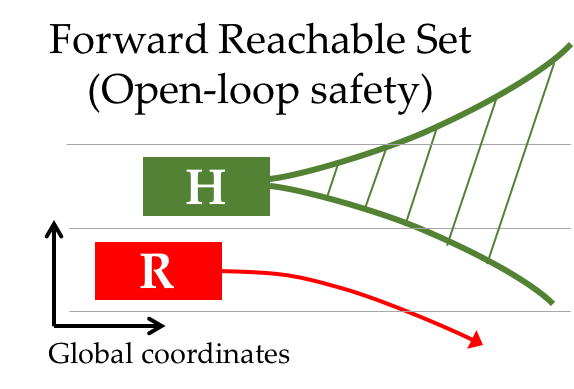}} 

\caption{Illustration of backward and forward reachable sets (shaded) to be avoided.}
\label{fig:reachability panel}
\end{figure}

\begin{figure}
    \centering
    \includegraphics[width=\textwidth]{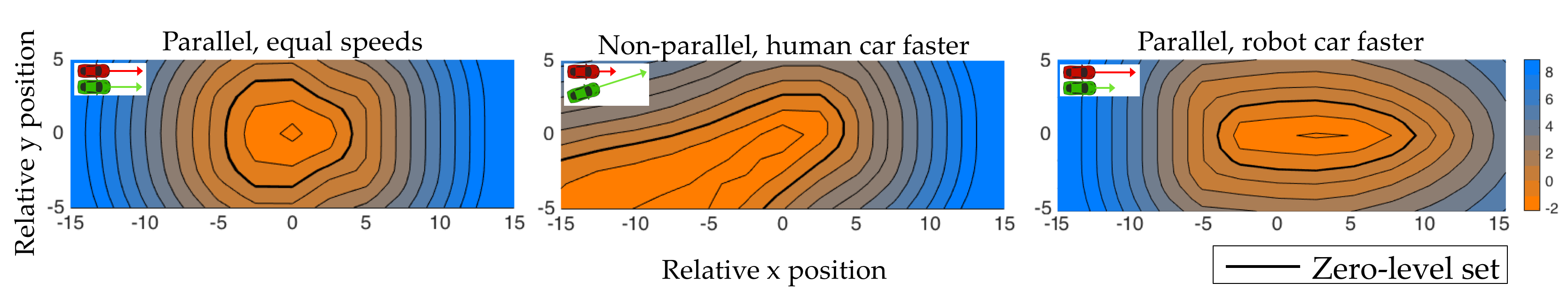}
    \caption{Contour plots of slices of the HJI value function for the relative dynamics~\eqref{eqn:relative dynamics}; slices show $V$ as a function of $\xrel$ and $\yrel$ with all other states held fixed. Left: The pear-shaped BRS stems from %
    the fact the robot car can swerve its front more rapidly than its rear, e.g., to avoid collision. Middle: If the human car is traveling faster (in this case at an angle), it is unsafe for the robot to be in front of the human. Right. If the robot car is traveling faster, it should not be directly behind the human as collision may be unavoidable if the human brakes abruptly.}
    \label{fig:collision_avoid_set}
\end{figure}

\subsection{Backward Reachability Analysis}
We use infinite time horizon backward reachability analysis to compute additional safety constraints for the MPC tracking controller.
Critically, \emph{backward reachability} is computed backwards in time allowing for closed-loop robot reactions to avoid future collisions (see Figure~\ref{fig:BRS illustration}), while \emph{forward reachability} is computed forwards in time assuming open-loop control for the robot (see Figure~\ref{fig:FRS illustration}) which leads to an overly-conservative robot outlook. Though more expensive to compute, backward reachability solutions may be precomputed offline and cached for online use. We briefly review relevant HJ reachability definitions in the remainder of this section; see \cite{ChenTomlin2018} for a more in-depth treatment.

Let the relative dynamics of the robot and human be given by $\dot{x}_{\rel} = f(x_{\rel}, u_R, u_H)$, and the collision set be $\mathcal{T}$. 
The BRS $\mathcal{A}(t)$ represents the set of ``avoid states'' from which if the human followed an adversarial strategy $u_H(\cdot)$, any robot action $u_R(\cdot)$ would lead to the relative state trajectory $x_{\rel}(\cdot)$ being inside $\mathcal{T}$ within a time horizon $|t|$ (note that $t<0$ when propagating backwards in time):
{\footnotesize 
\begin{equation}
    \mathcal{A}(t) := \lbrace  \bar x_{\rel} \in \mathbb{R}^n: \exists u_H(\cdot), \forall u_R (\cdot), \exists s\in[t, 0], x_\rel(t) = \bar x_\rel\wedge\dot x_{\rel} = f(x_{\rel}, u_R, u_H) \wedge x_{\rel}(s) \in \mathcal{T} \rbrace
    \label{eqn: backward reachable tube}
\end{equation}
}

\noindent Assuming optimal human actions, $\mathcal{A}(t)$ can be computed by solving the Hamilton-Jacobi-Isaacs (HJI) partial differential equation \cite{MitchellBayenEtAl2005,FisacChenEtAl2015}, whose solution $V(t, x_\rel)$ gives the BRS as its zero sublevel set: $\mathcal{A}(t) = \lbrace x_\rel : V(t,x_\rel) \leq 0\rbrace$.\footnote{The HJI PDE is solved starting from the boundary condition $V(0, x_\rel)$, the sign of which reflects set membership of $x_\rel$ in $\mathcal{T}$; see Section~\ref{sec:insights} for discussion of specific choices of $V(0, x_\rel)$.} When the control capabilities of the human are no greater than those of the robot, one can take the limit $t\rightarrow-\infty$ and obtain the infinite time horizon BRS $\mathcal{A}_\infty$ with corresponding value function $V_\infty(x_\rel)$.\footnote{For ease of notation going forward we will often write $V := V_\infty$.} Illustrative slices of the value function and the BRS for the vehicle-vehicle relative dynamics considered in this work are shown in Figure~\ref{fig:collision_avoid_set}.

The optimal robot avoidance control offers the greatest increase in $V$ assuming optimal (worst-case) actions by the human:

{\footnotesize
\begin{equation}
 u_R^* = \arg \max_{u_R} \min_{u_H}\nabla V(x_{\rel}) \cdot f(x_{\rel}, u_R, u_H)
 \label{eqn: optimal control}
\end{equation}
}

\noindent Previous applications of HJI solutions have switched to this control when near the boundary of the BRS \cite{FisacAkametaluEtAl2017}, i.e., when safety is nearly violated.
In an interactive scenario where, for example, we may want to let a robot planner convey intent by nudging towards the human car to the extent that is safe, we prefer a less extreme control strategy.
Rather than switching to the optimal avoidance controller \eqref{eqn: optimal control}, we add the set of safety-preserving controls defined below as an additional constraint in the MPC optimal control problem.

{\footnotesize
\begin{equation}
    \mathcal U_R(x_{\rel}) = \{u_R: \min_{u_H} \nabla V(x_{\rel}) \cdot f(x_{\rel}, u_R, u_H) \ge 0\}.
    \label{eq:safe preserve}
\end{equation}
}

\noindent $\mathcal U_R(x_\rel)$ represents the set of robot controls that ensure the value function is nondecreasing.
We compute the BRS offline using the BEACLS toolkit \cite{TanabeChen2018}, and cache the value function $V$ and its gradient $\nabla V$.
Online we employ a safety buffer $\epsilon > 0$ so that when the condition $V(x_{\rel}) \leq \epsilon$ holds, indicating that the robot is nearing safety violation, we add a half-space constraint $\mathcal U_R(x_{\rel})= \lbrace u_R:  M_{HJI}\cdot u_R+b_{HJI}\geq 0 \rbrace$ to the list of MPC constraints in the tracking control problem. This constraint is a linearized approximation of \eqref{eq:safe preserve}; specifically $M_{HJI} = \nabla V  \frac{\partial f}{\partial u_R}(x_\rel, u_H^*, u_R)$ and $b_{HJI} = \nabla V \cdot f(x_\rel, u_H^*, u_R) - M_{HJI}\cdot u_R$ evaluated at the current state and robot control ($u_H^*$ denotes the optimal, i.e., worst-case, human action defined analogously to~\eqref{eqn: optimal control}).

\section{Methodology}
In this section we describe how we tailor the technical approach described above to the particular application of ensuring safety for human-robot pairwise vehicle interactions. We detail the relative dynamics model for HJI computation and outline the MPC tracking controller; all the details of the traffic weaving interaction planner can be found in \cite{SchmerlingLeungEtAl2018}.
All code for the planner, MPC tracking controller, and BRS computation is available at \url{https://github.com/StanfordASL/safe_traffic_weaving}.

\subsection{Relative Dynamics}
We use a six-state single track model to describe the robot car's dynamics (see \cite{BrownFunkeEtAl2017} for details) and assume the human car obeys a dynamically extended unicycle model.
This represents a compromise between model fidelity and the number of state dimensions in the relative dynamics; reducing the latter is essential since solving the HJI PDE suffers greatly from the curse of dimensionality.\footnote{The computation becomes notoriously expensive past five or more states without compromising on grid discretization or employing some decoupling strategy. We use a grid size of $13\times 13\times 9\times 9\times 9\times 9\times 9$ for our 7D system uniformly spaced over $(\xrel, \yrel, \psirel, \Uxr, \Uyr, \vh, \rr) \in [-15, 15] \times [-5, 5] \times [-\pi/2, \pi/2] \times [1, 12] \times [-2, 2] \times [1, 12] \times [-1, 1]$; computing the BRS with this discretization takes approximately 16 hours on a 3.6GHz octocore AMD Ryzen 1800X CPU.}

We define the relative dynamics \eqref{eqn:relative dynamics} in a coordinate system centered on and aligned with the robot, where $R_\theta$ is the counterclockwise rotation matrix over $\theta$.
$\xrel$, $\yrel$ and $\psirel$ are the relative position and heading angle between the human and robot, while the other states are needed to completely describe the human and robot car dynamics. The robot control inputs $u_R = [\delta,\,\Fxf, \Fxr]$ are steering angle and front and rear longitudinal tire force (we write $F_x = \Fxf + \Fxr$ to denote total longitudinal force), while the human control inputs $u_H=[\omega,\,a]$ are yaw rate and longitudinal acceleration.
{\footnotesize
\begin{align}
\small
\begin{matrix}
\begin{bmatrix}
\xrel\\ \yrel 
\end{bmatrix} = R_{-\psir} \begin{bmatrix}
\xh - \xr \\ \yh - \yr
\end{bmatrix}\\
\\
\psirel = \psih - \psir
\end{matrix}  \qquad x_\rel = \begin{bmatrix}
\dxrel\\ \dyrel\\\dpsirel\\\dUxr\\\dUyr\\\dvh\\\drr
\end{bmatrix} = \begin{bmatrix}
\vh \cos \psirel - \Uxr + \yrel \rr\\
\vh \sin \psirel - \Uyr - \xrel \rr\\
\omega - \rr\\
\frac{1}{m} (\Fx + F_{x_{drag}}) - \rr \Uxr\\
\frac{1}{m}(\Fyf + \Fyr) - \rr\Uxr\\
a\\
\frac{1}{I_{zz}}(\bar{a}\Fyf - \bar{b}\Fyr)\\
\end{bmatrix}
\label{eqn:relative dynamics}
\end{align}
}

\noindent We assume that the robot and human car share the same power, steering, and friction limits. These limits are chosen to reflect the same physical constraints assumed for the MPC tracking controller's dynamic bicycle model \cite{BrownFunkeEtAl2017}. Due to its simpler dynamics representation, the human car has a transient advantage in control authority over the robot car (it may change its path curvature discontinuously, while the robot may not), but by equating the steady state control limits we ensure that the infinite time horizon BRS computation converges.

\subsection{The MPC+HJI Tracking Controller}
Both the trajectory tracking objective and safety-preserving control constraint rely on optimizing over the robot steering and longitudinal force inputs simultaneously. The associated tracking optimization problem is non-convex; we apply an approximate variant of sequential quadratic programming (SQP) in which we time-discretize and linearize the dynamics over a lookahead horizon and solve one corresponding QP at each MPC step. We interpolate along each solution trajectory to compute the linearization nodes for the QP at the next MPC step.\footnote{Only a single iteration of SQP is solved for the tracking problem at each MPC step, rather than the usual iteration until convergence. Since the tracking problems are so similar from one MPC step to the next, we find that this approach yields sufficient performance for our purposes.}

We use the \verb|ForwardDiff.jl| automatic differentiation (AD) package implemented in the Julia programming language \cite{RevelsLubinEtAl} to linearize the trajectory tracking dynamics as well as the HJI relative dynamics to facilitate the computation of $M_{HJI}$ and $b_{HJI}$. We call the Operator Splitting Quadratic Program (OSQP) solver \cite{StellatoBanjacEtAl2017} through the \verb|Parametron.jl| modeling framework \cite{Koolencontributors}; this combination of software enables us to solve the following MPC optimization problem at 100Hz.

Let $q_k = [\Delta s_k, U_{x,k}, U_{y,k}, r_k, \Delta \psi_k, e_k]^T$ be the state of the robot car with respect to a nominal trajectory at discrete time step $k$. $\Delta s_k$,  $e_k$ and $\Delta \psi_k$ denote longitudinal, lateral, and heading angle error; $U_{x,k}$, $U_{y,k}$, and $r_k$ are body-frame longitudinal and lateral velocity, and yaw rate respectively. Let $u_k=[\delta_k, F_{x,k}]$ be the controls at step $k$ and let $A_k q_k + B_k^- u_k + B_k^+ u_{k+1} + c_k = q_{k+1}$ denote linearized first-order-hold dynamics. We adopt the varying time steps method and stable handling envelope constraint ($H_k$, $G_k$) from \cite{BrownFunkeEtAl2017}; $\sigma_{\beta,k}$, $\sigma_{r,k}$, and $\sigma_{HJI}$ are slack variables to ensure the existence of a feasible solution. The constraint $M_{HJI} u_j + b_{HJI} \geq -\sigma_{HJI}$ is added only when $V(x_{\rel}) \leq \epsilon$. Although HJI theory suggests that applying this constraint on the next action alone is sufficient, we apply it over the next 3 timesteps (30ms lookahead) to account for the many approximations inherent in our QP formulation.

{\footnotesize
\begin{equation}
\begin{aligned}
\minimize_{q, u, \sigma, \sigma_{HJI}, \Delta\delta, \Delta F_x}  & \;\;\sum_{k=1}^T\Delta s_k ^T Q_{\Delta s} \Delta s_k + \Delta \psi_k ^T Q_{\Delta \psi} \Delta \psi_k + e_k ^T Q_e e_k  +  \Delta \delta_k ^T R_{\Delta  \delta}  \Delta \delta_k + \\
& \qquad \Delta F_{x,k}^T R_{\Delta \Fx} \Delta F_{x,k} + W_{\beta} \sigma_{\beta,k} + W_r\sigma_{r,k} + W_{HJI}\sigma_{HJI,k}\\
\text{subject to} \quad  & \delta_{k+1} - \delta_k = \Delta \delta_k, \quad \Delta \delta_{min} \leq \Delta \delta_k \leq \Delta \delta_{max}, \quad \delta_{min} \leq \delta_{k} \leq \delta_{max}\\
& F_{x,k+1} - F_{x,k} = \Delta F_{x,k}, \quad V_{min} \leq U_{x,k} \leq V_{min}, \quad F_{x,min} \leq F_{x,k} \leq F_{x,max}  \\
& \sigma_{1,k} \geq 0, \quad \sigma_{2,k} \geq 0 , \quad \sigma_{HJI,j} \geq 0\\
& H_k \begin{bmatrix} U_{y,k} \\ r_k \end{bmatrix} - G_k \leq \begin{bmatrix} \sigma_{\beta,k} \\ \sigma_{r,k} \end{bmatrix},\quad A_k q_k + B_k^- u_k + B_k^+ u_{k+1} + c_k = q_{k+1} \\
& q_1 = q_{curr}, \quad u_1=u_{curr}, \quad M_{HJI} u_j + b_{HJI} \geq -\sigma_{HJI} \\
&\text{for } j = 1,...,T_{HJI}, \quad k=1,...,T
\end{aligned}
\label{eqn:QP}
\end{equation}
}

\section{Experiments}
\subsection{Experimental Vehicle Platform}
\begin{figure}[t]
    \centering
    \includegraphics[width=\textwidth]{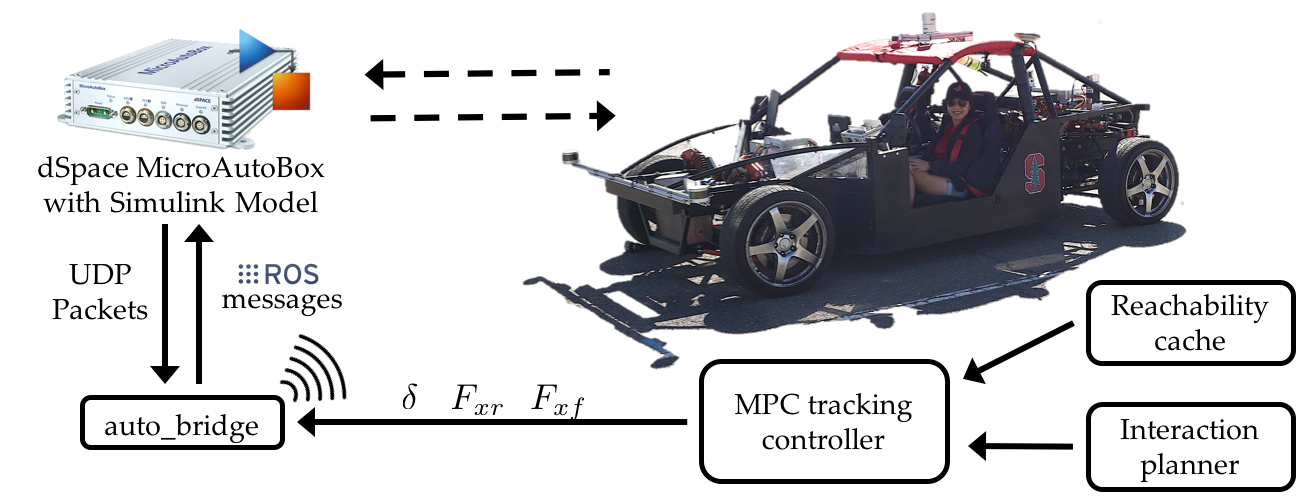}
    \caption{X1: a steer-by-wire experimental vehicle platform. It is equipped with three LiDARs (one 32-beam and two 16-beam), a differential GPS/INS which provides pose estimates accurate to within a few centimeters as well as high fidelity velocity, acceleration, and yaw rate estimates. ROS is used to interface the planning/control stack described in this work with a dSpace MicroAutoBox onboard X1 which handles sensing and control at the hardware level.}
    \label{fig:X1 label}
\end{figure}
X1 is a flexible steer-by-wire, drive-by-wire, and brake-by-wire experimental vehicle developed by the Stanford Dynamic Design Lab (see Figure \ref{fig:X1 label}). To control X1, desired steering ($\delta$) and longitudinal tire force ($\Fxf, \Fxr$) commands are sent to the dSpace MicroAutoBox (MAB) which handles all sensor inputs except LiDAR (handled by the onboard PC) and implements all low level actuator controllers.
We use the Robot Operating System (ROS) to communicate with the MAB; the planning/control stack described in this work is running onboard X1 on a consumer desktop PC running Ubuntu 16.04 equipped with a quadcore Intel Core i7-6700K CPU and an NVIDIA GeForce GTX 1080 GPU.

\subsection{Results}\label{sec:experiments}

\begin{wrapfigure}{r}{0.2\textwidth}
    \vspace{-.5cm}
    \includegraphics[width=\textwidth]{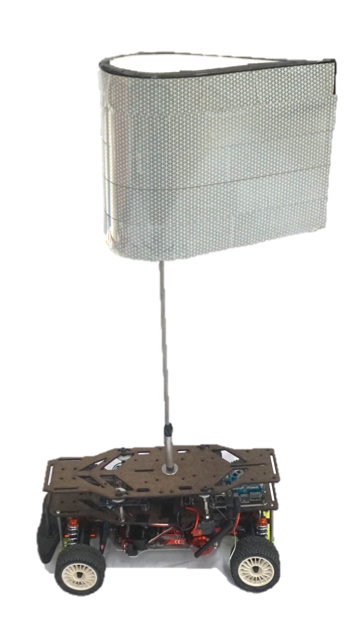}
    \caption{1/10-scale RC~car~with~LiDAR- visible mast.}
    \label{fig: rc car with lidar visible mast}
\end{wrapfigure}

To evaluate our proposed control stack --- a synthesis of a high-level probabilistic interaction planner with the MPC+HJI tracking controller --- we perform full-scale human-in-the-loop traffic-weaving trials with X1 taking on the role of the robot car.\footnote{We scaled the highway traffic-weaving scenario (mean speed $\sim$28m/s) in \cite{SchmerlingLeungEtAl2018} down to a mean speed of $\sim$8m/s by shortening the track (reducing longitudinal velocity by a constant) and scaling time by a factor of 4/3 (with the effect of scaling speeds by 3/4 and accelerations by 9/16).} 
We investigate and evaluate the effectiveness of our control stack by allowing the human car to act carelessly (i.e., swerving blindly towards the robot car) during the experiments. We compare our proposed controller (MPC+HJI) against a tracking-only MPC controller (MPC) and a controller that switches to the HJI optimal avoidance controller when near safety violation (switching). To ramp up towards testing with two full-scale vehicles in the near future, we investigate two types of human car: (1) a virtual human-driven car and (2) a 1/10-scale LiDAR-visible human-driven RC car.

\subsection{Virtual Human-Driven Vehicle}
\begin{figure}
\centering
\subfloat[Controller comparison on experimental data where the robot car uses a MPC+HJI controller.]{
    \label{fig:HJI+MPC comparison}
    \includegraphics[width=\textwidth]{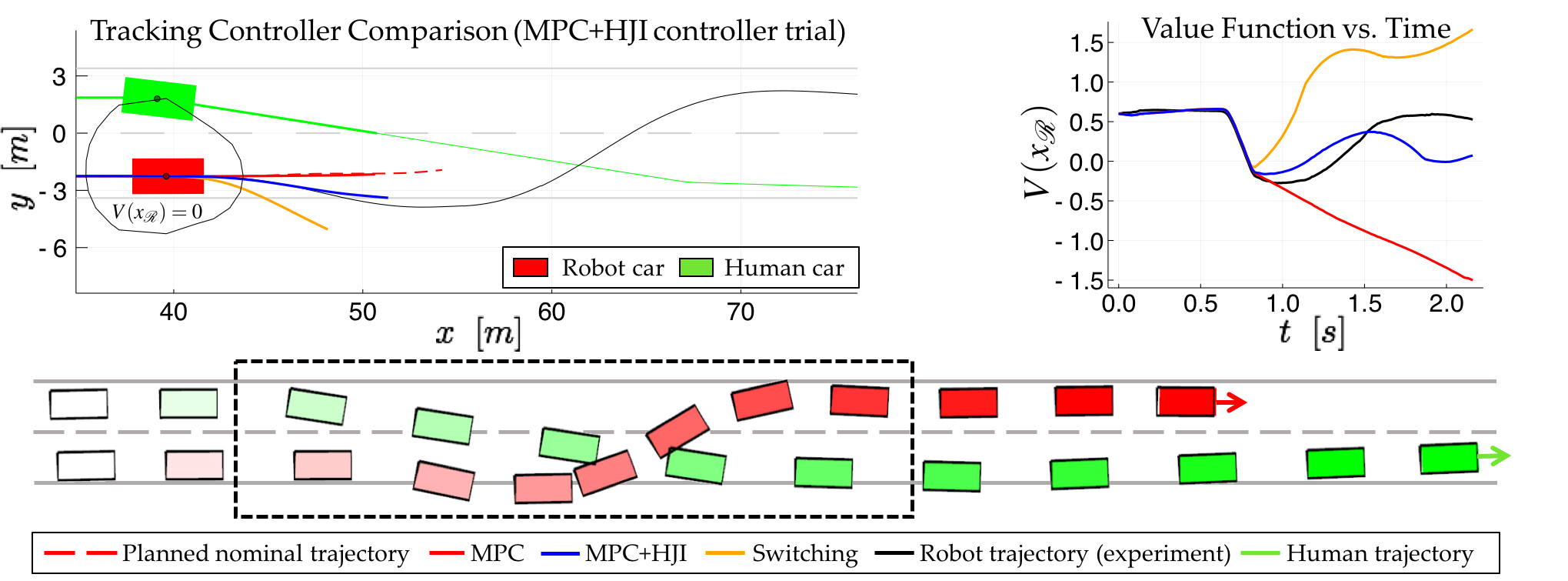} } 
 
\subfloat[Controller comparison on experimental data where the robot car uses a switching controller.]{
    \label{fig:switching comparison}
    \includegraphics[width=\textwidth]{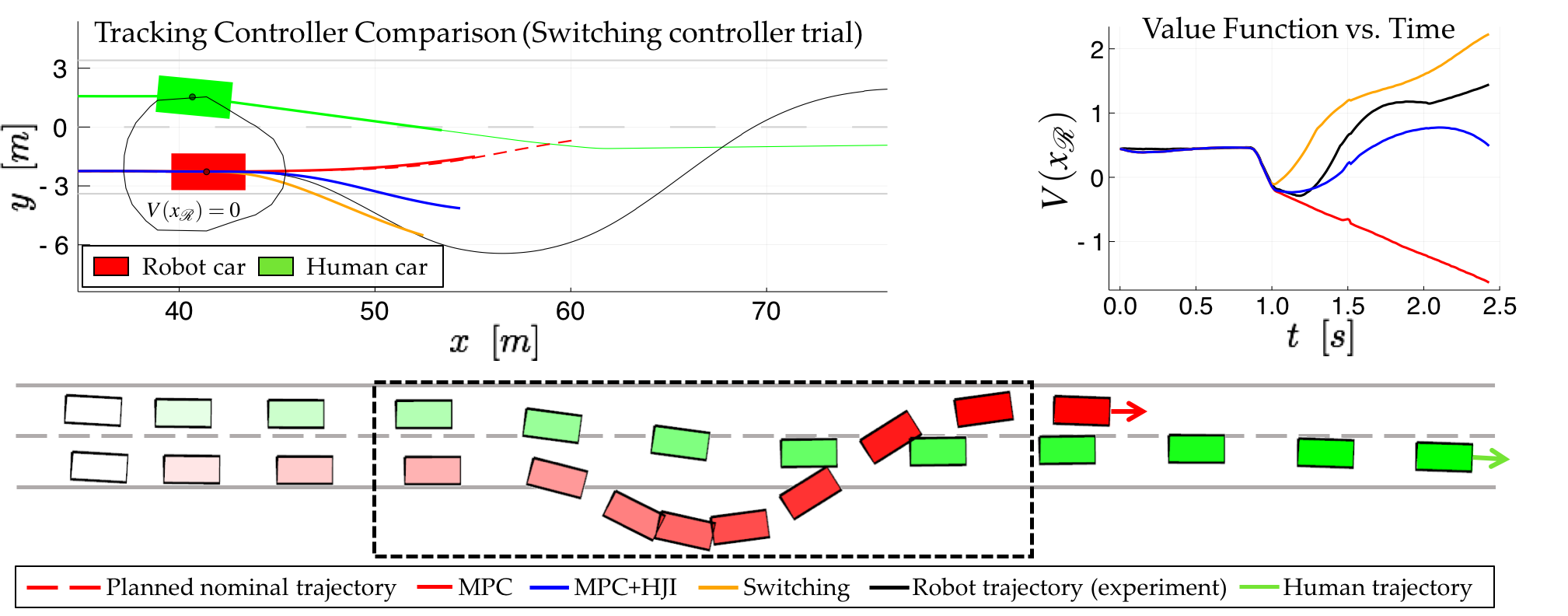}} 
\caption{Controller comparison on planner trajectories from X1/virtual human-driven vehicle experiments. Top left: A close up view of the different simulated controller trajectories when $V(x_\rel)$ first drops below $\epsilon = 0.05$. Top right: The corresponding evolutions of $V(x_\rel)$. Bottom: A time-lapse view of the traffic-weaving experiment, with the dashed box corresponding to top left figure.
}
\label{fig:controller comparison virtual car experiment}
\end{figure}

To ensure a completely safe experimental environment, our first tier of experimentation uses a joystick-controlled virtual vehicle for the human car and allows the robot control stack to have perfect observation of the human car state.
Experimental trials of the probabilistic planning framework for the MPC+HJI and switching controllers are shown in Figure~\ref{fig:HJI+MPC comparison} and \ref{fig:switching comparison} respectively, along with a simulated comparison between the two safety controllers and the tracking-only controller.\footnote{For comparative purposes, the controllers were simulated with the displayed nominal trajectory held fixed, but in reality, the nominal trajectory in these experiments was updated at $\sim 4Hz$.}
As expected, we see that the MPC+HJI controller represents a middle ground between the tracking-only MPC which does not react to the human car's intrusion, and the switching controller which arguably overreacts with a large excursion outside the lane boundaries. Evident from Figure~\ref{fig:HJI+MPC comparison}, our proposed controller tries to be minimally interventional --- the robot car swerves/brakes but only to an extent that is necessary.

\subsection{1/10-Scale Human-Driven Vehicle}
\begin{figure}
    \centering
    \includegraphics[width=\textwidth]{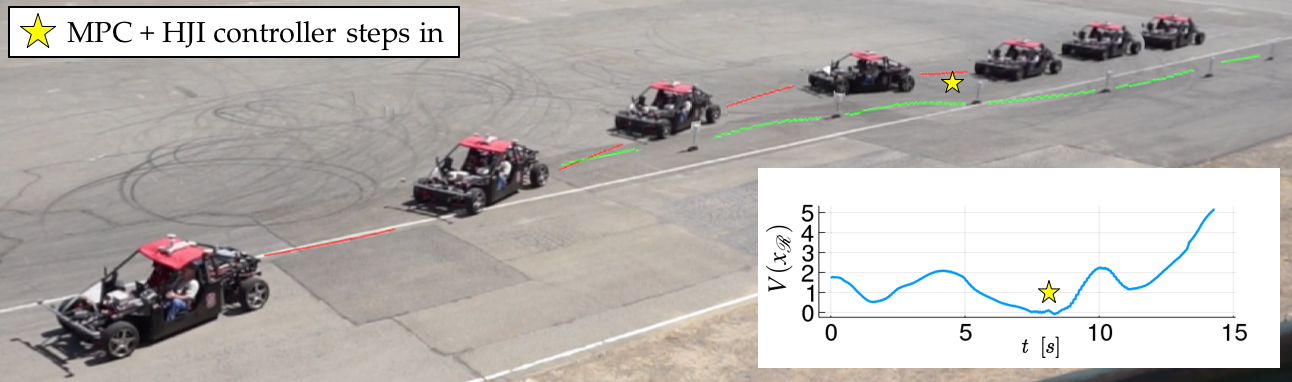}
    \caption{Time-lapse of pairwise vehicle interaction: X1 with 1/10-scale human-driven RC car. The RC car (green trajectory) nudges into X1 (red trajectory), which swerves gently to avoid.}
    \label{fig:rccar time lapse}
\end{figure}

To begin to investigate the effects of perception uncertainty on our safety assurance framework we use three LiDARs onboard X1 to track a human-driven RC car, and implement a Kalman filter for human car state estimation (position, velocity, and acceleration). Even with imperfect observations, we show some successful preliminary results (an example is shown in Figure~\ref{fig:rccar time lapse}) at mean speeds of ~4m/s, close to the limits of the RC car + LiDAR-visible mast in crosswinds at the test track. We observe similar behavior as in the virtual human car experiments, including the fact that the value function dips briefly below zero before the MPC+HJI controller is able to arrest its fall; we discuss this behavior in the next section.

\section{Discussion}\label{sec:insights}
Beyond the qualitative confirmation of our design goals, our experimental results reveal two main insights.

\vspace{.1cm}
\noindent \emph{Takeaway 1: The reachability cache is underly-conservative with respect to robot car dynamics and overly-conservative with respect to human car dynamics.}
\vspace{.1cm}

\noindent In all cases --- hardware experiments as well as simulation results --- the HJI value function $V$ dips below zero, indicating that neither the HJI+MPC nor even the optimal avoidance switching controller are capable of guaranteeing safety in the strictest sense. The root of this apparent paradox is in the computation of the reachability cache used by both controllers as the basis of their safety assurance. Though the 7-state relative dynamics model~\eqref{eqn:relative dynamics} subsumes a single-track vehicle model that has proven successful in predicting the evolution of highly dynamic vehicle maneuvers \cite{FunkeBrownEtAl2017,BrownFunkeEtAl2017}, the way it is employed in computing the value function $V$ omits relevant components of the dynamics. In particular, when computing the optimal avoidance control~\eqref{eqn: optimal control} as part of solving the HJI partial differential equation, we assume total freedom over the choice of robot steering angle $\delta$ and longitudinal force command $F_x$, %
up to maximum control limits. This does not account for, e.g., limits on the steering slew rate (traversing $[-\delta_{max}, \delta_{max}]$ takes approximately 1 second), and thus the value function is computed under the assumption that the robot can brake/swerve far faster than it actually can.

We note that simply tuning the safety buffer $\epsilon$ is insufficient to account for these unmodeled dynamics. In Figure~\ref{fig:controller comparison virtual car experiment} we see that $V$ may drop from approximately 0.5 (the value when the two cars traveling at 8m/s start side-by-side in lanes) to -0.3 in the span of a few tenths of a second. Selecting $\epsilon > 0.5$ might give enough time for the steering to catch up, but such a selection would prevent the robot car from accomplishing the traffic weaving task even under nominal conditions, i.e., when the human car is equally concerned about collision avoidance. This is because the safety controller would push the robot car outside of its lane from the outset to maintain the buffer. This behavior follows from wide level sets associated with the transient control authority asymmetry (recall that in the HJI relative dynamics the human car may adjust its trajectory curvature discontinuously), assumed as a conservative safety measure as well as a way to keep the relative state dimension manageable.

The simplest remedy for both of these issues is to increase the fidelity of the relative dynamics model by incorporating additional integrator states $\dot\delta$ for the robot and $\dot \omega$ for the human. Naively increasing the state dimension to 8 or 9, however, might not be computationally feasible (even offline) without devising more efficient HJI solution techniques or choosing an extremely coarse discretization over the additional states.\footnote{By literature standards we already use a relatively coarse discretization grid for solving the HJI PDE; associated numerical inaccuracies may be another source of the observed safety mismatch.} We believe that simulation, accounting for slew rates, could be a good tool to prototype such efforts, noting that as it stands we have relatively good agreement between simulation and experimental platform in our testing.

\vspace{.1cm}
\noindent\emph{Takeaway 2: Interpretability of the value function $V$ should be a key consideration in future work.}
\vspace{.1cm}

\noindent In this work the terminal value function $V(0, x_\rel)$ is specified as the separation/pene-tration distance between the bounding boxes of the two vehicles, a purely geometric quantity dependent only on $\xrel$, $\yrel$, and $\psirel$. Recalling that V ($:= V_\infty$) represents the worst-case eventual outcome of a differential game assuming optimal actions from both robot and human, we may interpret the above results through the lens of worst-case outcomes, i.e., a value of -0.3 may be thought of as 30cm of collision penetration assuming optimal collision seeking/avoidance from human/robot. %
When extending this work to cases with environmental obstacles (e.g., concrete highway boundaries that preclude large deviations from lane), or multi-agent settings where the robot must account for the uncertainty in multiple other parties' actions, for many common scenarios it may be the case that guaranteeing absolute safety is impossible. Instead of avoiding a BRS of states that might lead to collision, we should instead treat the value function inside the BRS as a cost. In particular we should specify more contextually relevant values of $V(0, x_\rel)$ for states in collision, e.g., negative kinetic energy or another notion of collision severity as a function of the velocity states $\Uxr$, $\Uyr$, $\vh$, and $\rr$ in addition to the relative pose. This would lead to a controller that prefers, in the worst case, collisions at lower speed, or perhaps ``glancing blows'' where the velocities of the two cars are similar in magnitude and direction. Computationally we could use such a value function in an MPC formulation similar to that presented in this work, with static obstacles either incorporated in $V(0, x_\rel)$ or represented as additional MPC constraints as done in \cite{BrownFunkeEtAl2017}, and accommodating multiple other agents by considering them pairwise with the robot and taking the minimum over the corresponding value functions.

\section{Conclusions}
We have investigated a control scheme for providing real-time safety assurance to underpin the guidance of a probabilistic planner for human-robot vehicle-vehicle interactions. By essentially projecting the planner's desired trajectory into the set of safety-preserving controls whenever safety is threatened, we preserve more of the planner's intent than would be achieved by adopting the optimal control with respect to separation distance. Our experiments show that with our proposed minimally interventional safety controller, we accomplish the high level objective (traffic weaving) despite the human car swerving directly onto the path of the robot car, and accomplish this relatively smoothly compared to using a switching controller that results in the robot car swerving more violently off the road. We note that this work represents only a promising first step towards the integration of reachability-based safety guarantees into a probabilistic planning framework. We have already discussed the concrete modifications to this controller we believe are necessary to improve the impact of these guarantees; further study should also consider better fitting of the planning objective at the controller level. That is, instead of performing a naive projection, i.e., the one that minimizes trajectory tracking error, it is likely that a more nuanced selection informed by the planner's prediction model would represent a better ``backup choice'' in the case that safety is threatened. We recognize that ultimately, guaranteeing absolute safety on a crowded roadway may not be realistic, but we believe that in such situations value functions derived from reachability may provide a useful metric for near-instantly evaluating the future implications of a present action choice.

\bibliographystyle{IEEEtran-short}
{
\small
\bibliography{../../../../bib/main,../../../../bib/ASL_papers}
}

\end{document}